\documentclass[letterpaper,10pt,conference]{ieeeconf}

\IEEEoverridecommandlockouts
\usepackage{amsmath}
\usepackage{amssymb}
\usepackage{amsfonts}
\usepackage{graphicx}
\usepackage{xspace}
\usepackage{caption}
\usepackage{algorithm}
\usepackage{algpseudocode}
\usepackage{booktabs}
\usepackage[T1]{fontenc}
\usepackage{url}

\makeatletter
\newcommand\fs@spaceruled{%
  \def\@fs@cfont{\bfseries}\let\@fs@capt\floatc@ruled
  \def\@fs@pre{\vspace*{6pt}\hrule height.8pt depth0pt \kern2pt}%
  \def\@fs@post{\kern2pt\hrule\relax}%
  \def\@fs@mid{\kern2pt\hrule\kern2pt}%
  \let\@fs@iftopcapt\iftrue}
\makeatother

\floatstyle{spaceruled}
\restylefloat{algorithm}
\title{\LARGE \bf
SafeLoop: Risk-Aware Rollback for Vision-Language-Action Manipulation
}

\author{Zeyu Lou$^{1,\dagger}$, Tianran Zhang$^{2,\dagger}$, Xinquan Yue$^{1}$, Ya Jing$^{3}$, and Chenyang Si$^{1,*}$%
\thanks{$^{\dagger}$Equal contribution. $^{*}$Corresponding author.}%
\thanks{$^{1}$Zeyu Lou, Xinquan Yue, and Chenyang Si are with Nanjing University, Nanjing, China.
E-mails: {\tt\small zeyu.lou.mail@gmail.com}; {\tt\small chenyang.si@nju.edu.cn}}%
\thanks{$^{2}$Tianran Zhang is with The Hong Kong University of Science and Technology (Guangzhou), Guangdong, China.}%
\thanks{$^{3}$Ya Jing is with Beijing University of Technology, Beijing, China.}%
\thanks{\copyright\ 2026 IEEE. Personal use of this material is permitted. Permission from IEEE must be obtained for all other uses, in any current or future media, including reprinting/republishing this material for advertising or promotional purposes, creating new collective works, for resale or redistribution to servers or lists, or reuse of any copyrighted component of this work in other works.}%
}

\begin{document}
\maketitle
\begin{abstract}
Recent vision-language-action (VLA) models are promising for general-purpose manipulation, but long-horizon execution remains fragile. Small state-estimation or control errors can lead to irreversible failures (e.g., collisions and object drops). Avoiding these risks requires a proactive safety mechanism capable of anticipating hazards. In this paper, we introduce \textsc{SafeLoop}, a non-invasive external wrapper that adds hazard prediction and rollback-based recovery to a VLA model without changing its parameters. \textsc{SafeLoop} trains a risk predictor from vision and proprioception to output four values: the probability and time-to-hazard for body collisions and for object failures. A lightweight controller then chooses one of three actions based on the predicted risk: continue execution (\texttt{noop}), save a safety checkpoint (\texttt{record}), or retreat in joint space (\texttt{rollback}). Rollback moves the robot back to a recent safe waypoint and queries the base policy again, which may yield an alternative continuation. Across 24 LIBERO tasks (16 random seeds each) and three real-robot tasks (25 rollouts each), \textsc{SafeLoop} achieves a stronger overall safety--success trade-off than alternative methods, reducing hazard cases by roughly 70\% while preserving task success and the base-policy control rate. Project code is available at \url{https://github.com/Loule0-0/SafeLoop/tree/release/safeloop}.
\end{abstract}

\section{Introduction}
\label{sec:intro}

Recent foundation models have shown strong abilities in visual perception, semantic understanding, and reasoning \cite{beyer2024paligemmaversatile3bvlm,yang2025qwen3technicalreport,grattafiori2024llama3herdmodels}. These advances have naturally extended to robotics and led to vision-language-action (VLA) models, which are a promising path toward general-purpose manipulation \cite{kim2024openvla,zitkovich2023rt}. Recent works have shown that VLA models can map images and language instructions to actions, enabling robots to solve a wide range of tasks \cite{intelligence2025pi,gr00tn1_2025}.

\begin{figure}[t]
  \centering
  \includegraphics[width=\linewidth]{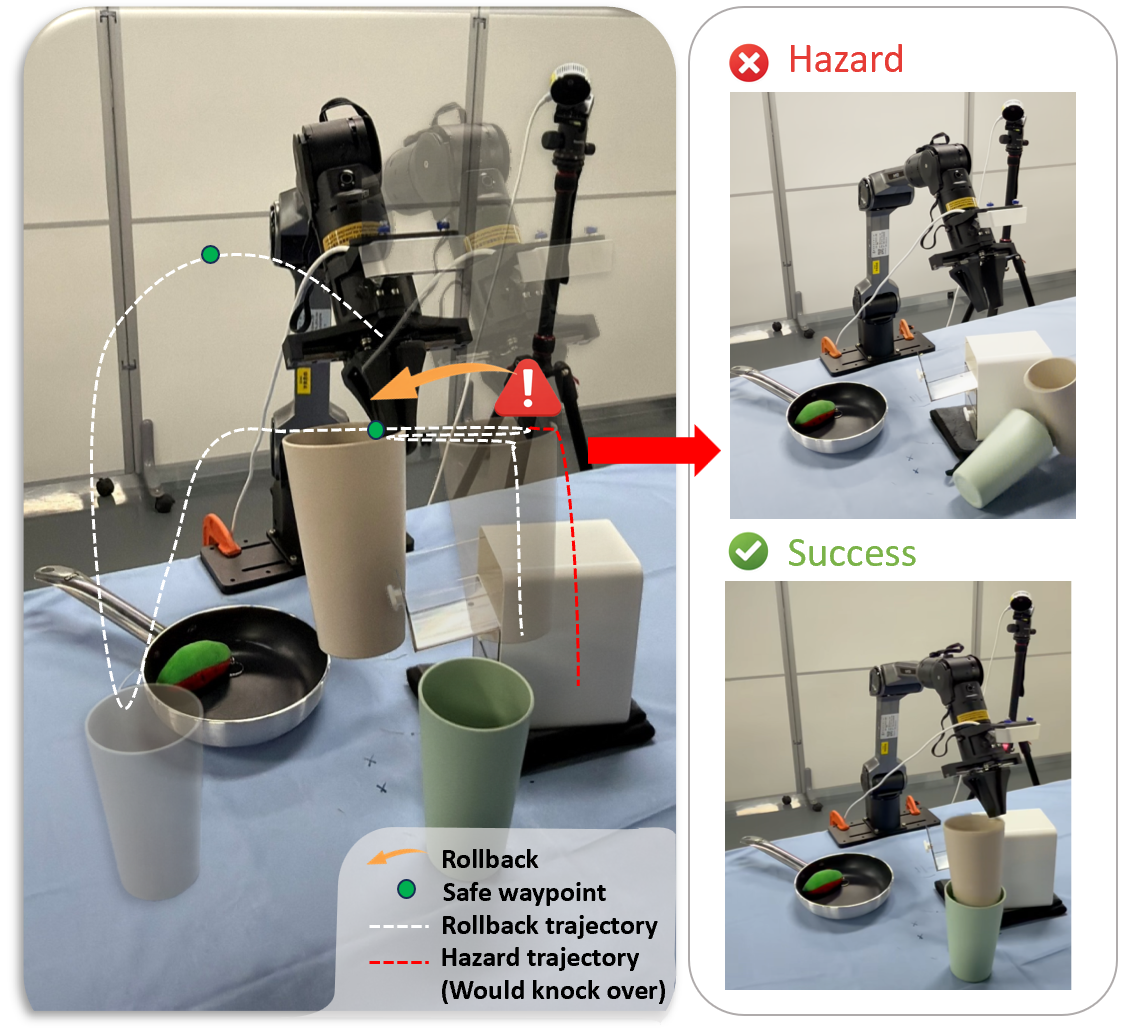}
  \caption{\textbf{\textsc{SafeLoop} rollback on a real robot.} During execution, the system dynamically saves safe waypoints (green dots). When the risk predictor flags an impending hazard (warning triangle), the decider triggers a \texttt{rollback} to the most recently saved waypoint. The policy is then queried again; in this example, the new continuation avoids the predicted collision (red dashed line) and completes the task (white dashed line).}
  \label{fig:rollback_teaser}
  \vspace{-1.0em}
\end{figure}

VLA models remain fragile in long-horizon manipulation because their action-generation paradigm offers limited ability to correct execution drift \cite{NEURIPS2024_ca92ff06}. Small state-estimation and control errors can persist across successive action chunks, compound over multiple stages, and eventually cause collisions, object failures, or task failure \cite{xue2025reactive}. Existing interventions therefore face deployment trade-offs at VLA scale. Runtime enforcement--formal shielding \cite{alshiekh2018shielding} or analytic action-projection layers \cite{dalal2018safe}--requires explicit constraints and safety-relevant state, which can be difficult to obtain in complex, vision-based manipulation. Training-time methods modify the policy or its optimization procedure \cite{zhang2025safevla,thananjeyan2021recovery,hsu2022improving,ak2023learning}, whereas predictive failure monitors require a separate fallback response \cite{farid2022failure}. Deploying frozen foundation models therefore calls for a proactive execution-time safety system that anticipates imminent hazards and guides the robot to a recoverable state.

\begin{figure*}[t]
  \vspace{2mm}
  \centering
  \includegraphics[width=\textwidth]{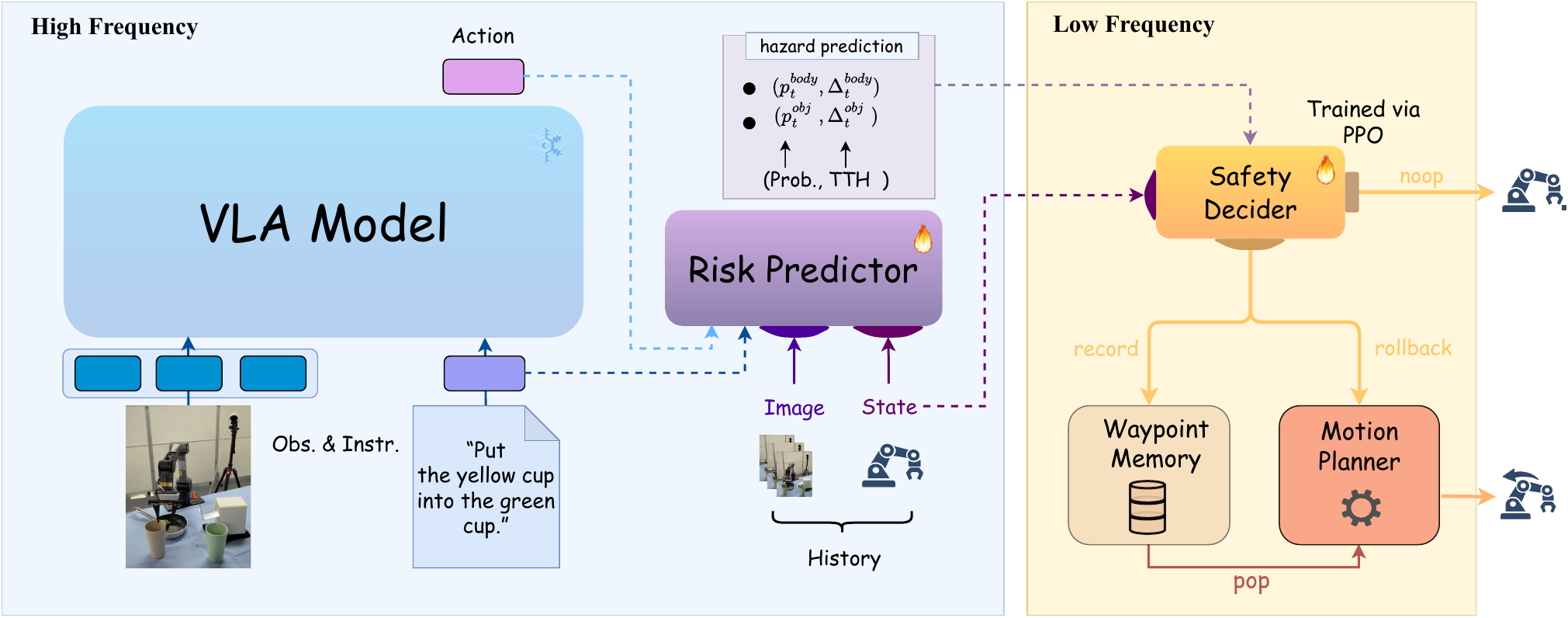}
  \caption{\textbf{Method overview.} A frozen policy proposes high-frequency actions, while an asynchronous risk predictor evaluates impending body- and object-level hazards and outputs their probabilities and time-to-hazard (TTH). At a lower frequency, a safety decider selects one of three interventions: (1) \texttt{noop} to proceed nominally, (2) \texttt{record} to store the current safe waypoint, or (3) \texttt{rollback} to execute a motion-planned recovery to a stored waypoint.}
  \label{fig:overview}
  \vspace{-3mm}
\end{figure*}

Realizing such a proactive safety system over a frozen VLA introduces three deployment challenges: \textbf{(i) Hazards are rare and varied.} Unsafe events do not happen often, may appear only after a delay, and come from different sources. Some hazards are robot-level, such as arm or base collisions, while others are task-level, such as drops or loss of grasp. A practical predictor therefore needs to estimate both how likely a hazard is and how soon it may occur. \textbf{(ii) Safety requires recovery, not just stopping.} Long-horizon manipulation cannot rely on stop-only rules. When risk is rising, the system should actively move away from the developing failure, return to a safe configuration, and then continue the task. Without a recovery path, even conservative stopping will often reduce success rate and leave the robot stuck. \textbf{(iii) VLA control is often stochastic.} Many modern VLA policies generate actions by sampling, so the same observation can lead to different future motions. This creates both a challenge and an opportunity. The safety layer must decide when to intervene and where to resume so that the policy can sample an alternative continuation, while still running in real time and working across different backbones.

To move beyond reactive halting, we introduce predictive safety for VLAs via \textsc{SafeLoop}, a non-invasive wrapper that equips frozen policies with hazard forecasting and proactive kinematic recovery without altering their parameters. Upon anticipating an imminent hazard (Fig.~\ref{fig:rollback_teaser}), \textsc{SafeLoop} preempts the nominal action and triggers a \texttt{rollback} to a historical safe waypoint before resuming execution.

Operating asynchronously on the real system (Fig.~\ref{fig:overview}), an onboard risk predictor forecasts continuous hazard probabilities and time-to-hazard. A low-frequency decider maps this risk vector to discrete interventions: continuing (\texttt{noop}), checkpointing a safe waypoint (\texttt{record}), or recovering (\texttt{rollback}). By restoring a prior kinematic configuration, rollback lets the stochastic base policy sample an alternative continuation that may avoid the developing hazard.

Across different VLA backbones, \textsc{SafeLoop} reduces hazard rates by 56--75\% (e.g., OpenVLA-OFT body-hazard rate drops from 22.9\% to 5.7\%) and decreases hazard events per 1,000 steps by roughly 80\% (1.03$\rightarrow$0.21), while maintaining task success (67.7\% vs. 68.2\%). Because \textsc{SafeLoop} does not alter the base policy's task competence, non-hazard task failures remain; accordingly, its safety gains are larger than the change in success rate.

{\looseness=-1
Our contributions are threefold: (1) a non-invasive, dual-rate safety wrapper for frozen VLA policies; (2) a four-output predictor that forecasts body- and object-level hazard probabilities and TTH from onboard observations; and (3) a learned three-action decider with safe-waypoint rollback, validated across three VLA backbones in simulation and on physical robots without real-world RL training of the decider.\par}

\section{Related Work}
\label{sec:related_work}

\noindent\textbf{Vision-Language-Action Models.}
Transformer-based robot policies model multimodal observations and action sequences for long-horizon manipulation \cite{brohan2022rt,zhao2023learning}. VLAs further reuse vision-language representations to map images and instructions to actions and improve transfer across tasks \cite{kim2024openvla,zitkovich2023rt}. Diffusion-style generators provide expressive action-sequence models \cite{chi2025diffusion,Ze2024DP3}, while large-scale robot pretraining improves generalist behavior \cite{black2024pi_0,lbmtri2025}. Recent systems additionally study high-level semantic prediction and action world models for longer-horizon coherence \cite{intelligence2025pi_,cen2025worldvla,cen2025rynnvla}. Most of this literature primarily optimizes task success and generalization rather than hazard anticipation and execution-time recovery.

\noindent\textbf{Safety in manipulation, and safety mechanisms for VLAs.}
Safety in robot control has long relied on model-based mechanisms, including reactive collision avoidance \cite{khatib1986real}, trajectory optimization with obstacle costs \cite{zucker2013chomp,schulman2013finding}, robust MPC tracking under uncertainty \cite{nubert2020safe}, and operational-space safety control with torque saturation \cite{murtaza2021real}. Learning-based variants add hierarchical safety behaviors \cite{ak2023learning}, shielding with reachability-style certificates \cite{thumm2022provably}, and generative control conditioned on safety costs \cite{deng2025safebimanual}; perception and semantics can also modulate interaction safety \cite{amaya2022vision,brunke2025semantically}. Closest to our setting are methods that wrap or modify VLAs directly: SafeVLA uses constrained learning during training \cite{zhang2025safevla}; VLSA (AEGIS) adds a control-barrier-function layer \cite{hu2025vlsa}; and SAFE predicts failures from internal VLA features to trigger stopping or backtracking \cite{gu2025safe}. In contrast, \textsc{SafeLoop} leaves the base policy frozen and couples external hazard forecasting with proactive recovery.

\noindent\textbf{Recovery and episodic safety memory.}
Recovery has been studied via backup controllers near constraint boundaries \cite{thananjeyan2021recovery}, hierarchical or multi-component frameworks with explicit verification and recovery mechanisms \cite{triantafyllidis2023hybrid,yang2025agentic}, and recovery policies that return the system to states from which nominal control can resume \cite{vats2024recoverychaining}. Memory-augmented methods store recovery experience for reuse \cite{hsu2022improving} or learn memory-writing and forgetting policies with explicit retrieval \cite{kim2023machine}; predictive monitors can forecast failure likelihood to trigger downstream intervention \cite{farid2022failure}. \textsc{SafeLoop} adapts these concepts to frozen VLAs by coupling predictive hazard estimates with episodic safe-waypoint memory and rollback without weight updates.

\section{Method}
\label{sec:method}

\textsc{SafeLoop} is a safety wrapper designed to prevent developing hazards from becoming irreversible, without retraining the VLA policy. Our method follows three phases:

\begin{enumerate}
    \item[1)] \textbf{Data collection and annotation.} We collect execution rollouts under the frozen $\pi_{\text{base}}$ and explicitly annotate safety-critical events to serve as hazard supervision.
    \item[2)] \textbf{Risk predictor training.} We distill these annotations into a short-horizon risk predictor. This lightweight model forecasts imminent hazards relying solely on onboard observations and the base policy's proposed actions.
    \item[3)] \textbf{Meta-policy training.} We train a low-frequency meta-policy via reinforcement learning. This controller maps continuous risk estimates into \texttt{noop}, \texttt{record}, or \texttt{rollback}, triggering physical recovery only when necessary.
\end{enumerate}

\begin{algorithm}[t]
\caption{\textsc{SafeLoop} Dual-Rate Control Loop}
\label{alg:safeloop}
\begin{algorithmic}[1]
\Require Base policy $\pi_{\text{base}}$, risk predictor $g_\phi$, meta-policy $\pi_\theta$, period $K$
\State Init memory $\mathcal{M} \gets \emptyset$, timestep $t \gets 0$
\While{episode not terminated}
  \State Observe state $o_t$, robot configuration $q_t$, and instruction $c$
  \State $a_t \sim \pi_{\text{base}}(\cdot \mid o_t,c)$ \Comment{Nominal action}
  \State $\mathbf{r}_t \gets g_\phi(o_{\leq t},a_t,c)$ \Comment{Async risk eval}
  \If{$t \pmod K = 0$} \Comment{Low-freq arbitration}
    \State Build meta-state $\mathbf{x}_t$ from history
    \State Sample intervention $u_t \sim \pi_\theta(\cdot \mid \mathbf{x}_t)$
    \If{$u_t = \texttt{record}$}
      \State $\mathcal{M} \gets \mathcal{M} \cup \{(q_t,\mathbf{r}_t,t)\}$ \Comment{Save safe waypoint}
    \ElsIf{$u_t = \texttt{rollback}$}
      \State $m^* \gets \textsc{SelectSafeWaypoint}(\mathcal{M},t)$
      \State $\textsc{MemoryRollback}(m^*)$ \Comment{Recover}
      \State \textbf{continue} \Comment{Resume from waypoint}
    \EndIf
  \EndIf
  \State Execute $a_t$ in environment; $t \gets t+1$
\EndWhile
\end{algorithmic}
\end{algorithm}

Algorithm~\ref{alg:safeloop} summarizes the deployment-time control loop. The frozen base policy proposes actions at every low-level step, while an asynchronous risk predictor produces a compact risk vector for the current proposed action. A low-frequency decider reads recent risk/motion summaries to select an intervention, and a safety arbiter multiplexes between nominal actions and rollback control. Here $t$ indexes nominal base-policy steps: rollback is executed by a separate recovery controller and advances physical time without consuming another nominal VLA step.

\subsection{Hazard Formulation and Risk Prediction}
\label{subsec:hazard_data}
\label{subsec:risk_head}

\begin{figure}[t]
  \centering
  \includegraphics[width=\linewidth]{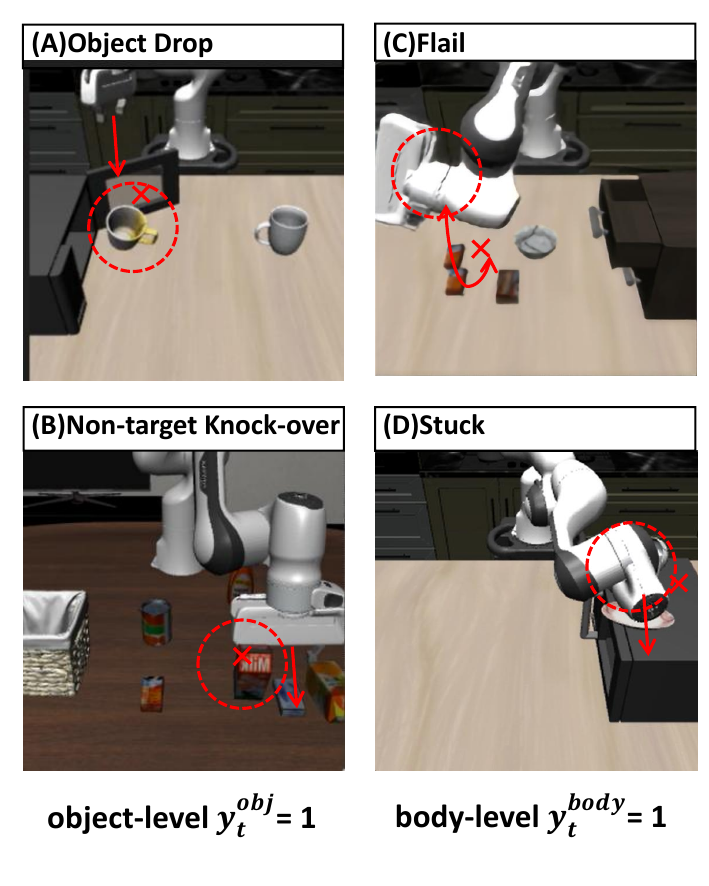}
  \caption{\textbf{Hazard taxonomy.} \emph{Object-level} hazards (left) include object drops (A) and non-target knock-overs (B). \emph{Body-level} hazards (right) include uncontrolled flailing (C) and the arm getting stuck or wedged (D).}
  \label{fig:hazard_examples}
\end{figure}

\begin{figure}[t]
  \vspace{2mm}
  \centering
  \includegraphics[width=\linewidth]{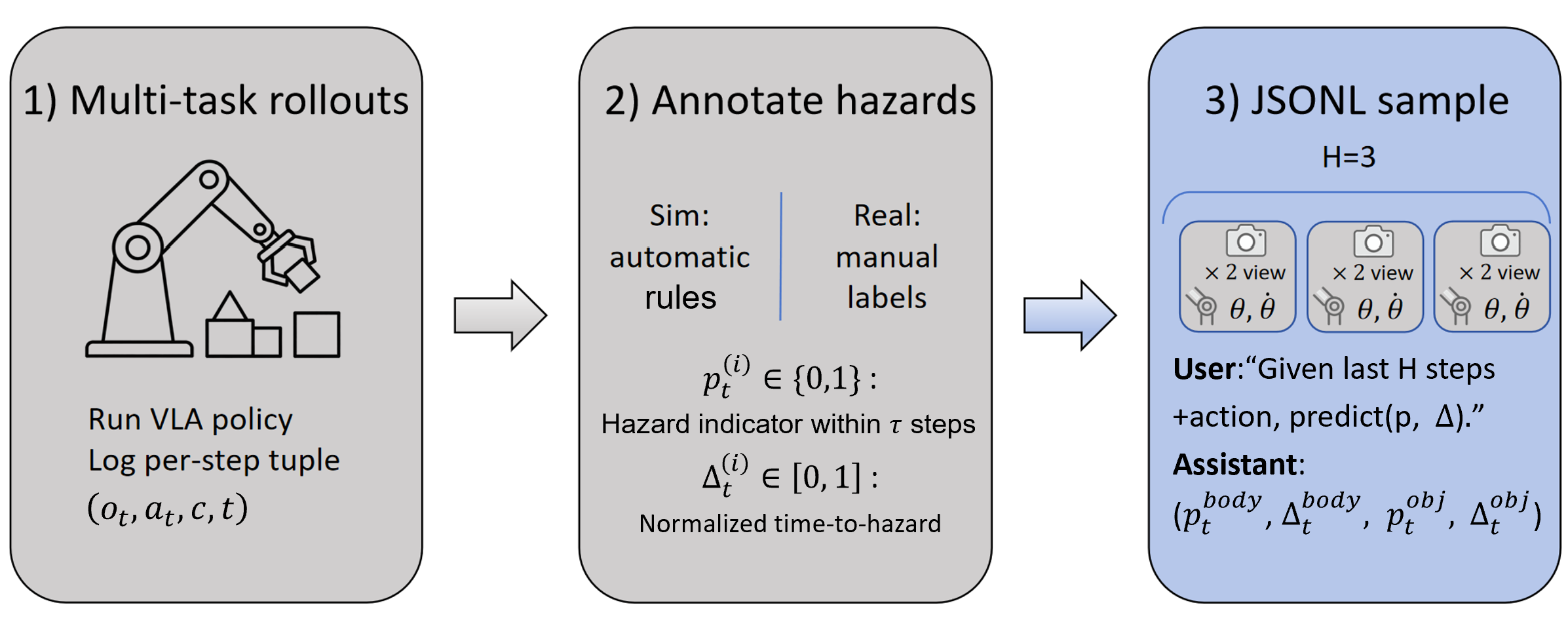}
  \caption{\textbf{Data pipeline for risk fine-tuning.} Raw rollouts are first annotated for hazards via simulation oracles or human labels. Spatial-temporal streams (images, proprioception, actions) are then segmented into fixed-length windows to compute short-horizon risk targets. Note the final serialization step, which formats these windows into instruction-style pairs to supervise the multimodal risk head.}
  \label{fig:data_to_template}
\end{figure}

We define a \emph{hazard} as a safety-critical event risking damage or loss of control, distinct from an ordinary task failure (e.g., a missed grasp). We model two types (Fig.~\ref{fig:hazard_examples}): \textbf{body-level hazards} $y_t^{\mathrm{body}}\in\{0,1\}$ include collisions, stuck states, and uncontrolled motion; \textbf{object-level hazards} $y_t^{\mathrm{obj}}\in\{0,1\}$ include drops, topples, and non-target disturbance. Training labels come from simulator-only contacts, object states, and kinematics through rule-based detectors, or from manually annotated real-robot clips. Kinematics only construct simulator labels for contact, object instability, and lack of end-effector progress; they are not deployment inputs, while object hazards require visual reasoning beyond inverse kinematics. The streams are segmented into fixed-length windows (Fig.~\ref{fig:data_to_template}).

Because hazards are rare, we combine regular and denser pre-hazard windows. Two auxiliary current-state classifiers share the backbone to label newly collected trajectories, but are not part of the four-dimensional forecast. Data needs scale mainly with hazard diversity: new types require representative labels, whereas the decider is reusable when the four risk semantics remain calibrated.

Let $o_t=(I_t,s_t)$ denote onboard observations (multi-view RGB $I_t$ and proprioception $s_t$), and let the frozen base policy propose low-level actions
\begin{equation}
a_t \sim \pi_{\text{base}}(\cdot \mid o_t,c),
\end{equation}
given instruction $c$. For each hazard $\mathrm{type}\in\{\mathrm{body},\mathrm{obj}\}$, define the first future hazard time
\begin{equation}
t_*^{(\mathrm{type})}(t)=\inf\{t'\geq t:y_{t'}^{(\mathrm{type})}=1\},
\end{equation}
with $t_*^{(\mathrm{type})}(t)=\infty$ if a hazard never occurs. Given a prediction horizon of $\tau$ low-level steps, we define two training targets:
\begin{align}
p_t^{(\mathrm{type})} &= \mathbb{I}\!\left\{t_*^{(\mathrm{type})}(t)\leq t+\tau\right\}, \\
\Delta_t^{(\mathrm{type})} &= \frac{1}{\tau}\min\!\left(t_*^{(\mathrm{type})}(t)-t,\tau\right),
\end{align}
where $p_t^{(\mathrm{type})}\in\{0,1\}$ indicates whether a hazard occurs within the next $\tau$ steps (the model output $\hat p$ is interpreted as probability), and $\Delta_t^{(\mathrm{type})}\in[0,1]$ is normalized time-to-hazard (smaller means more imminent; if $t_*=\infty$, then $\Delta=1$).

The risk predictor $g_\phi$ maps the sensor history $\{(I_{t-h},s_{t-h})\}_{h=0}^{H-1}$, current instruction $c$, and proposed action $a_t$ to the four-dimensional forecast
\begin{equation}
\mathbf{r}_t=
\big[\hat p_t^b,\hat\Delta_t^b,\hat p_t^o,\hat\Delta_t^o\big]^\top
=g_\phi\!\left(\{(I_{t-h},s_{t-h})\}_{h=0}^{H-1},a_t,c\right),
\end{equation}
where $i\in\{b,o\}$ denotes body- and object-level hazards. This vector is the safety forecast consumed by the decider. We train it with
\begin{equation}
\mathcal{L}_{\mathrm{risk}}=\sum_t\sum_{i\in\{b,o\}}
\left[\mathcal{L}_{\mathrm{BCE}}\!\left(p_t^i,\hat p_t^i\right)
+\lambda_\Delta\left\|\Delta_t^i-\hat\Delta_t^i\right\|_1\right].
\end{equation}

\subsection{Safety Decider and Asymmetric RL Optimization}
\label{subsec:decider}

At runtime, the low-frequency decider intervenes at step $t_k$ by selecting a discrete action $u_k\in\{\texttt{noop},\allowbreak\texttt{record},\allowbreak\texttt{rollback}\}$ based on recent joint motion and risk summaries. We formalize this sequential decision-making process as a discrete-action Markov Decision Process (MDP).

\textbf{MDP formulation.}
To bypass the reality gap while leveraging exact simulation states, we adopt an asymmetric actor--critic formulation. The actor policy observes deployable signals by aggregating recent joint configurations and predicted risk:
\begin{equation}
\mathbf{x}_k^{\mathrm{actor}}=\big[\mathbf{q}_{t_j},\mathbf{r}_{t_j}\big]_{j=k-L+1}^{k}.
\end{equation}
Conversely, the value function (critic) operates strictly during simulation-based training, utilizing augmented privileged states $\mathbf{z}_t$ (e.g., exact contact forces and rigid-body kinematics):
\begin{equation}
\mathbf{x}_k^{\mathrm{critic}}=\big[\mathbf{x}_k^{\mathrm{actor}},[\mathbf{z}_{t_j}]_{j=k-L+1}^{k}\big].
\end{equation}

We use a shaped reward that trades off task progress, hazard avoidance, and intervention economy:
\begin{equation}
r_k=r_k^{\mathrm{task}}+\lambda_{\mathrm{hz}}r_k^{\mathrm{hazard}}+\lambda_{\mathrm{int}}r_k^{\mathrm{intervene}},
\end{equation}
where $r_k^{\mathrm{hazard}}$ penalizes oracle-detected hazards (training-time only) and $r_k^{\mathrm{intervene}}$ penalizes unnecessary \texttt{record} / \texttt{rollback}. Exact coefficients are provided in the released configuration.

\textbf{Learning with PPO and a rule-based warm-start.}
Purely reward-driven learning can be brittle early on because hazards are sparse and recovery requires coordinated timing. We therefore design a heuristic teacher policy $\pi_{\mathrm{rule}}(\cdot\mid\mathbf{x}_k)$ that maps predicted risk and urgency to discrete interventions while respecting record/rollback cooldowns. It triggers \texttt{rollback} at high risk, triggers \texttt{record} below a conservative risk margin, and defaults to \texttt{noop} otherwise. The teacher provides an optimization warm start and is never queried at deployment.

To bootstrap the RL process, we initially warm-start the decider via Behavioral Cloning (BC) against this heuristic:
\begin{equation}
\mathcal{L}_{\mathrm{BC}}(\theta)=\mathbb{E}\!\left[\mathrm{CE}\!\left(\pi_{\mathrm{rule}}(\cdot\mid\mathbf{x}_k),\pi_\theta(\cdot\mid\mathbf{x}_k^{\mathrm{actor}})\right)\right].
\end{equation}

To ensure a smooth transition from imitation to pure reinforcement learning, we optimize the decider with PPO \cite{schulman2017proximal} augmented by a decaying BC objective. Let $\theta_{\mathrm{old}}$ denote the behavior policy
\begin{equation}
\rho_k(\theta)=\frac{\pi_\theta(u_k\mid\mathbf{x}_k^{\mathrm{actor}})}{\pi_{\theta_{\mathrm{old}}}(u_k\mid\mathbf{x}_k^{\mathrm{actor}})}.
\end{equation}
Using advantage estimates $\hat A_k$ computed from the asymmetric value function $V_\psi(\mathbf{x}_k^{\mathrm{critic}})$, the clipped surrogate loss is
\begin{equation}
\mathcal{L}_{\mathrm{PPO}}(\theta)=-\mathbb{E}_k\!\left[\min\!\left(\rho_k\hat A_k,\operatorname{clip}(\rho_k,1-\epsilon,1+\epsilon)\hat A_k\right)\right].
\end{equation}
The decider is optimized end-to-end minimizing the unified objective
\begin{equation}
\mathcal{L}_{\mathrm{total}}(\theta)=\mathcal{L}_{\mathrm{PPO}}(\theta)+\lambda_{\mathrm{BC}}\mathcal{L}_{\mathrm{BC}}(\theta),
\end{equation}
where the imitation coefficient $\lambda_{\mathrm{BC}}$ is linearly annealed to zero over the course of training.

\subsection{Implementation Details}
\label{subsec:implementation}

\textbf{Runtime and models.}
On hardware, \textsc{SafeLoop} runs the risk predictor asynchronously while retaining the base-policy control rate; simulation invokes it synchronously at decision boundaries, which is equivalent under the discrete arbitration schedule. We train a compact head on a frozen Qwen2.5-VL-3B backbone, use $H=3$ frames and a 20-step decision interval (approximately 1~Hz at 20~Hz control), and implement the decider as a width-128 MLP.

\textbf{Actor and critic observations.}
The actor receives a 49-D vector: at each of $L=3$ history steps, a 9-D joint/gripper state is concatenated with the four values in $\mathbf{r}_t$ and two auxiliary current-state probabilities, followed by four scalars for record validity, rollback availability, memory occupancy, and episode progress. The TTH entries are normalized to $[0,1]$, with smaller values indicating more imminent hazards. The auxiliary probabilities share the predictor backbone but remain separate from the four-output forecast. Seven simulator-only hazard and outcome signals are appended for the training-time critic, yielding 56 dimensions; the exported actor never receives them.

\textbf{Recovery and training.}
A \texttt{record} action stores the joint configuration and predicted risk; a \emph{safe waypoint} denotes a checkpoint below the configured risk threshold, not a formal certificate. Rollback selects the lowest-risk mature waypoint and plans a collision-checked joint path, rejecting the intervention if no valid target or path exists. Simulation uses joint interpolation with state restoration only as fallback, whereas hardware uses a low-level joint controller. The PPO reward favors completion, penalizes hazards and unnecessary interventions, and assigns outcome credit to rollback; exact weights are in the released configuration. The BC coefficient starts at 1.0 and is annealed to zero over the first 80\% of updates, so the threshold-based teacher only warm-starts optimization and is absent at deployment. Cooldowns and a rollback budget limit immediate cycles, but deterministic failures may repeat and irreversible changes cannot be undone.

\section{Experiments}
\label{sec:experiments}

In this section, we address four questions: \textbf{(I)} Can \textsc{SafeLoop} reduce hazards across VLA backbones while preserving task success? (\S~\ref{subsec:sim_results}); \textbf{(II)} Does its safety mechanism generalize to policy-OOD tasks and outperform baselines? (\S~\ref{subsec:ood_and_prior}); \textbf{(III)} Is rollback essential compared with simpler alternatives? (\S~\ref{subsec:ablations}); and \textbf{(IV)} Can the decider transfer to real robots without real-world RL? (\S~\ref{subsec:real_robot}).

\subsection{Experimental Setup}
\label{subsec:exp_setup}

\textbf{Task suites.} The simulation risk predictor is trained from annotated rollouts drawn from all four standard LIBERO suites (\textsc{Object}, \textsc{Goal}, \textsc{10}, and \textsc{Spatial}) \cite{liu2023libero}, whereas the decider is trained only on \textsc{Libero\_10}. For evaluation, we sample 24 tasks from these suites and \textsc{Libero\_90}. This setting tests whether the \textsc{Libero\_10}-trained decider transfers across task families when paired with a predictor trained across suites. We emphasize \textsc{Libero\_90} as a policy out-of-distribution (OOD) testbed: its tasks are unseen during base-policy task fine-tuning and introduce new layouts and contact patterns.

\textbf{Base policies.} We wrap three VLA controllers: \textsc{Pi-0} \cite{black2024pi_0}, \textsc{OpenVLA-OFT} \cite{kim2025openvla_oft}, and \textsc{Pi-0.5} \cite{intelligence2025pi_}. All methods use the same nominal-policy horizon and base-policy step limits.

\textbf{Evaluation protocol.} In simulation, we evaluate 24 tasks with 16 random seeds each (384 rollouts per method) and report macro-averages over tasks. On the real robot, we run 25 rollouts per task. The protocol does not assume that the scene is fully reversible: rollback restores the robot configuration, whereas irreversible object changes remain hazards or failures. Unless otherwise specified, all reported numbers are means over seeds or rollouts.

\textbf{Baselines.} We compare \textsc{SafeLoop} against two safety baselines: (1) RL: a safety decider trained directly via reinforcement learning to avoid hazards without an explicit risk predictor; and (2) SAFE \cite{gu2025safe}: a VLA failure-detection approach, evaluated with both reactive stopping (\texttt{+ stop}) and backtracking (\texttt{+ backtrack}).

\textbf{Ablations.} To isolate the efficacy of memory-based rollback and learned decision-making, we evaluate: (1) Resample, which replaces rollback with a new action sampled at the current state; (2) Rule-based Teacher, which replaces the PPO-trained decider with fixed risk thresholds; (3) Stop-on-Hazard, which halts on risk detection (omitted from the main plots due to near-zero success); and (4) Predictor Input variants that remove visual or temporal information (No Image, Context $H{=}1$).

\textbf{Metrics.} We report Success Rate and three safety metrics. Haz-Body is the percentage of rollouts containing at least one body-level hazard (collision, stuck state, or uncontrolled motion), and Haz-Obj is the percentage containing at least one object-level hazard (drop, topple, or unintended non-target disturbance). Events/1k counts transitions from safe to hazardous states per 1,000 nominal policy steps, measuring distinct occurrences rather than hazard duration.

\subsection{Success at Scale on VLA Backbones}
\label{subsec:sim_results}

\begin{figure}[t]
  \vspace{2mm}
  \centering
  \includegraphics[width=\linewidth]{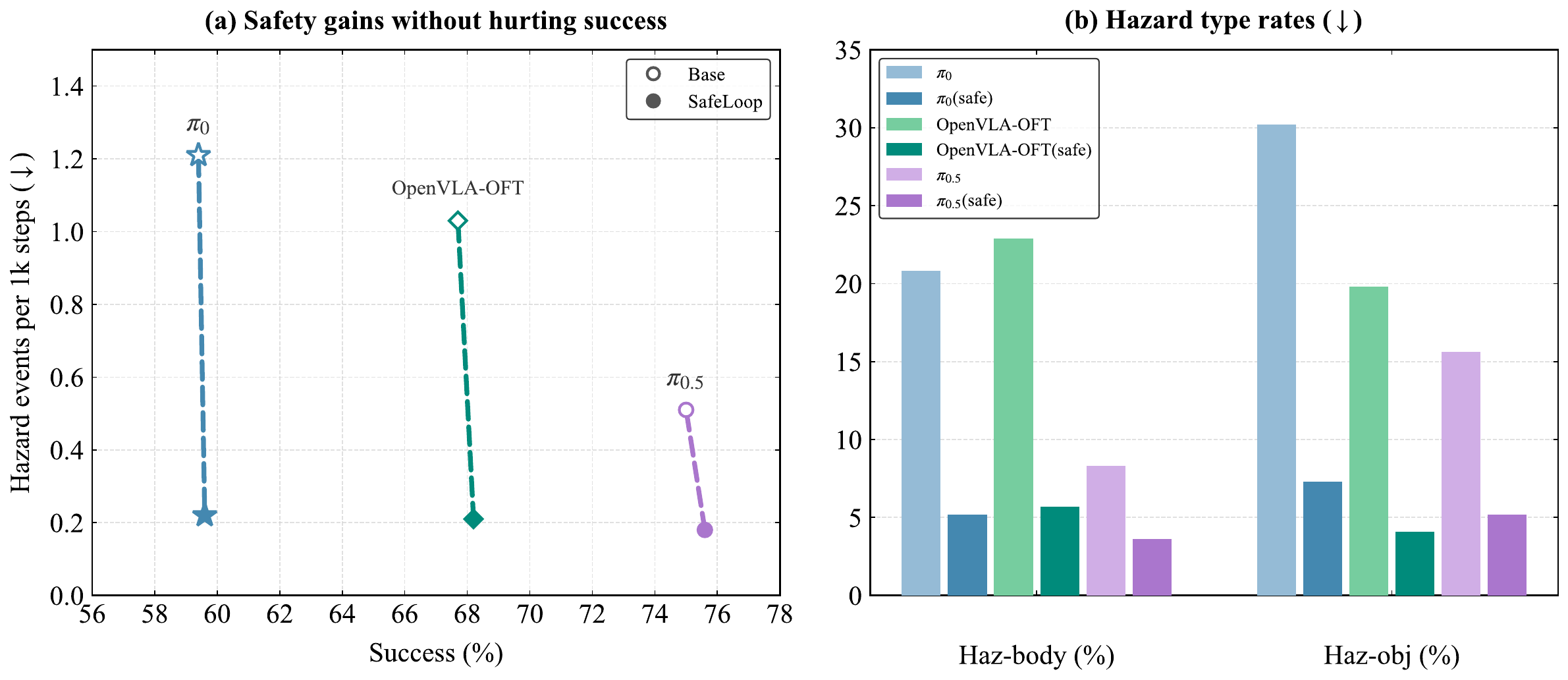}
  \caption{\textbf{Simulation results (macro-avg over 24 tasks).} (a) Safety--success trade-off: \textsc{SafeLoop} (solid markers) reduces hazard events across three VLA backbones relative to base policies (hollow markers) while maintaining comparable success. (b) Hazard breakdown: \textsc{SafeLoop} (filled bars) consistently suppresses both body- and object-level hazards compared with the baselines (outlined bars).}
  \label{fig:sim_results}
\end{figure}

\textbf{Non-invasive safety improvements across backbones.}
We apply \textsc{SafeLoop} as a wrapper around three VLAs and evaluate the safety--success trade-off in Fig.~\ref{fig:sim_results}. Across all backbones, \textsc{SafeLoop} sharply reduces hazards while maintaining comparable task success. For \textsc{OpenVLA-OFT}, body-level hazards drop from $22.9\% \to 5.7\%$ with a slight improvement in success ($67.7\% \to 68.2\%$). \textsc{SafeLoop} also reduces body-level hazards on the stronger \textsc{Pi-0.5} model ($8.3\% \to 3.6\%$). The drop in Events/1k (e.g., $1.03 \to 0.21$ for \textsc{OpenVLA-OFT}) indicates fewer repeated hazard events rather than early termination alone. Because the wrapper does not improve the frozen VLA's semantic task competence, unrelated task failures remain and the success-rate change is correspondingly smaller than the safety gain.

\subsection{Comparative Performance: \textsc{SafeLoop} vs. Other Methods}
\label{subsec:ood_and_prior}

\begin{table}[t]
  \vspace{2mm}
  \small
  \setlength{\tabcolsep}{4pt}
  \renewcommand{\arraystretch}{1.05}
  \centering
  \begin{tabular}{lcccc}
    \toprule
    Method & Success $\uparrow$ & Haz-B $\downarrow$ & Haz-O $\downarrow$ & Evts/1k $\downarrow$ \\
    \midrule
    \texttt{RL} & 53.9 & 7.0 & 9.1 & 0.32 \\
    \texttt{SAFE} + stop & 56.3 & 4.7 & 10.7 & 0.25 \\
    \texttt{SAFE} + backtrack & 58.0 & \textbf{4.4} & 9.9 & 0.24 \\
    \midrule
    \textbf{\textsc{SafeLoop}} & \textbf{59.6} & 5.2 & \textbf{6.8} & \textbf{0.22} \\
    \bottomrule
  \end{tabular}
  \vspace{4pt}
  \captionsetup{justification=raggedright,singlelinecheck=false}
  \caption{\textbf{Comparison to safety baselines.} \texttt{Haz-B} and \texttt{Haz-O} are rollout-level body- and object-hazard rates. \textsc{SafeLoop} achieves the highest success and lowest Haz-O and Events/1k; SAFE+backtrack obtains the lowest Haz-B.}
  \label{tab:safe_compare}
\end{table}

\textbf{Comparison to \texttt{SAFE} \cite{gu2025safe}}
As shown in Table~\ref{tab:safe_compare}, \texttt{SAFE} effectively predicts body-level hazards before they occur. However, it responds later to object-level hazards; risk scores often rise only after an object has been dropped. A halting strategy (\texttt{SAFE} + stop) can terminate episodes prematurely, while \texttt{SAFE} + backtrack incurs additional steps and can exceed the horizon before completion. \textsc{SafeLoop} instead forecasts both hazard types and uses its rollback mechanism to return to a recorded waypoint, reducing timeout failures.

\textbf{Generalization limits of end-to-end RL.}
The end-to-end \texttt{RL} baseline performs well on \textsc{Libero\_10} but degrades on policy-OOD \textsc{Libero\_90}; learning from high-dimensional histories is more sensitive to the training distribution. \textsc{SafeLoop} instead maps new scenes to four risk semantics consumed by a compact decider. The \textsc{Libero\_10}-trained decider is reused on unseen tasks and, after adapting only the predictor, on hardware without real-world RL.

\subsection{Ablation Studies: Impact of Key \textsc{SafeLoop} Design Choices}
\label{subsec:ablations}

\begin{table}[t]
  \vspace{2mm}
  \small
  \setlength{\tabcolsep}{4pt}
  \renewcommand{\arraystretch}{1.05}
  \centering
  \resizebox{\linewidth}{!}{%
  \begin{tabular}{lcccc}
    \toprule
    Method & Success $\uparrow$ & Haz-Body $\downarrow$ & Haz-Obj $\downarrow$ & Evts/1k $\downarrow$ \\
    \midrule
    \textsc{$\pi_0$} (Base) & 59.4 & 20.8 & 30.2 & 1.21 \\
    \midrule
    \texttt{No Image} & 42.7 & 14.6 & 21.9 & 0.60 \\
    \texttt{Context $H{=}1$} & 43.7 & 16.1 & 19.8 & 0.61 \\
    \texttt{Resample} & \textbf{60.4} & 16.6 & 13.5 & 0.53 \\
    \texttt{Teacher} & 58.3 & 7.3 & 8.9 & 0.25 \\
    \midrule
    \textbf{\textsc{SafeLoop}} & 59.6 & \textbf{5.2} & \textbf{6.8} & \textbf{0.22} \\
    \bottomrule
  \end{tabular}}
  \captionsetup{justification=raggedright,singlelinecheck=false}
  \caption{\textbf{Ablation study on the $\pi_0$ base policy.} \textsc{SafeLoop} achieves the optimal safety--success trade-off, validating our architectural design choices for risk prediction and recovery.}
  \label{tab:ablation}
\end{table}

We use \textsc{Pi-0} as the base policy (Table~\ref{tab:ablation}), isolating which components are responsible for safety gains.

\textbf{Rollback outperforms static resampling.}
The \texttt{Resample} baseline---drawing new stochastic actions from the current state---maintains high success ($60.4\%$) but increases hazards compared to \textsc{SafeLoop} ($16.6\%$ Haz-Body vs. $5.2\%$; $0.53$ vs. $0.22$ Evts/1k). This gap exposes a flaw in static resampling: once the robot enters a \emph{hazardous basin of attraction} (e.g., wedged geometries or contact-induced oscillations), a new action cannot guarantee escape. Rollback resets the kinematics, physically extracting the robot from the hazardous local minimum to a stable prior state to resume the task.

\textbf{Temporal and visual context drive proactive prediction.}
Removing visual input (\texttt{No Image}) drops success to $42.7\%$ and increases hazards, showing proprioception alone cannot anticipate object interactions before contact. Truncating history to a single frame (\texttt{Context $H{=}1$}) degrades both success and safety. Risk prediction requires \emph{approach dynamics} (e.g., velocity toward obstacles), not static snapshots.

\textbf{Learned decision-making outperforms fixed thresholds.}
The heuristic \texttt{Teacher} reduces hazards but lowers success compared with the learned decider. Fixed thresholds trigger rollback prematurely in uncertain states, accumulating recovery costs over long horizons. BC uses this teacher only to stabilize early PPO updates; the deployed learned decider then balances the risk of continuing against the cost of intervening. Thus, the \texttt{Teacher} row evaluates fixed-threshold deployment rather than serving as a no-BC ablation.

\subsection{Real-Robot Transfer and Deployment Overhead}
\label{subsec:real_robot}

\begin{figure}[t]
  \vspace{2mm}
  \centering
  \includegraphics[width=\linewidth]{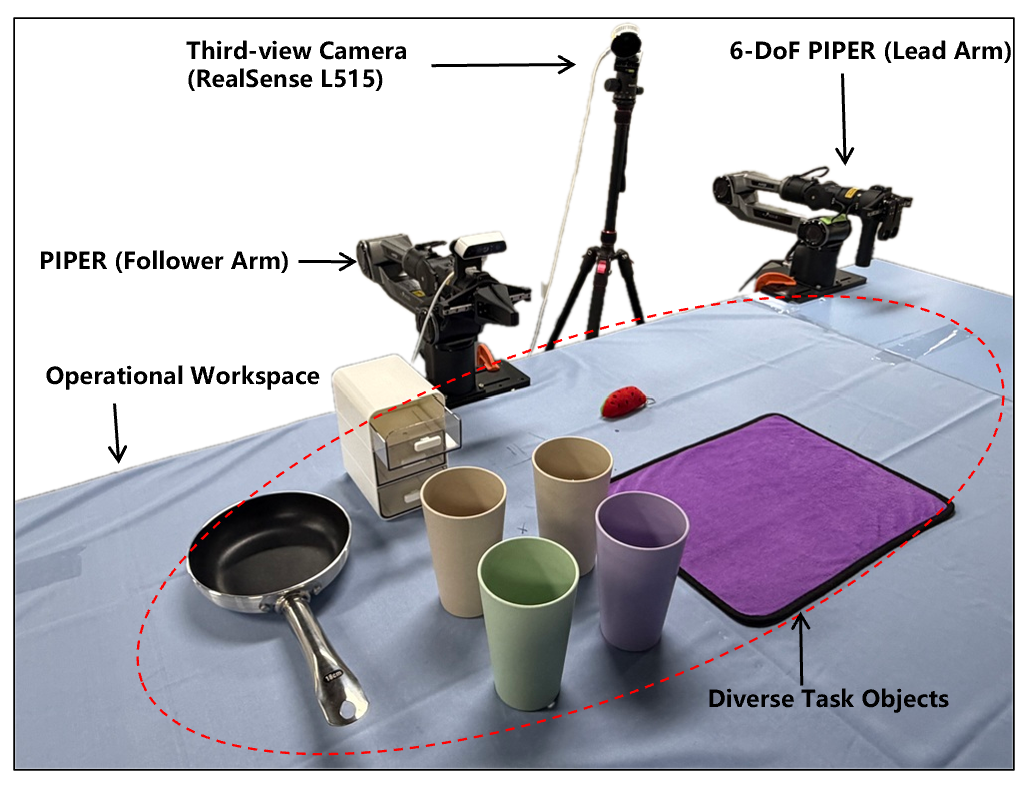}
  \caption{\textbf{Real-robot setup.} The physical evaluation platform comprises two 6-DoF PIPER arms and a third-view RealSense L515 camera. The workspace features diverse household objects with varying shapes, materials, and physical affordances.}
  \label{fig:real_setup}
\end{figure}

\begin{figure}[t]
  \centering
  \includegraphics[width=\linewidth]{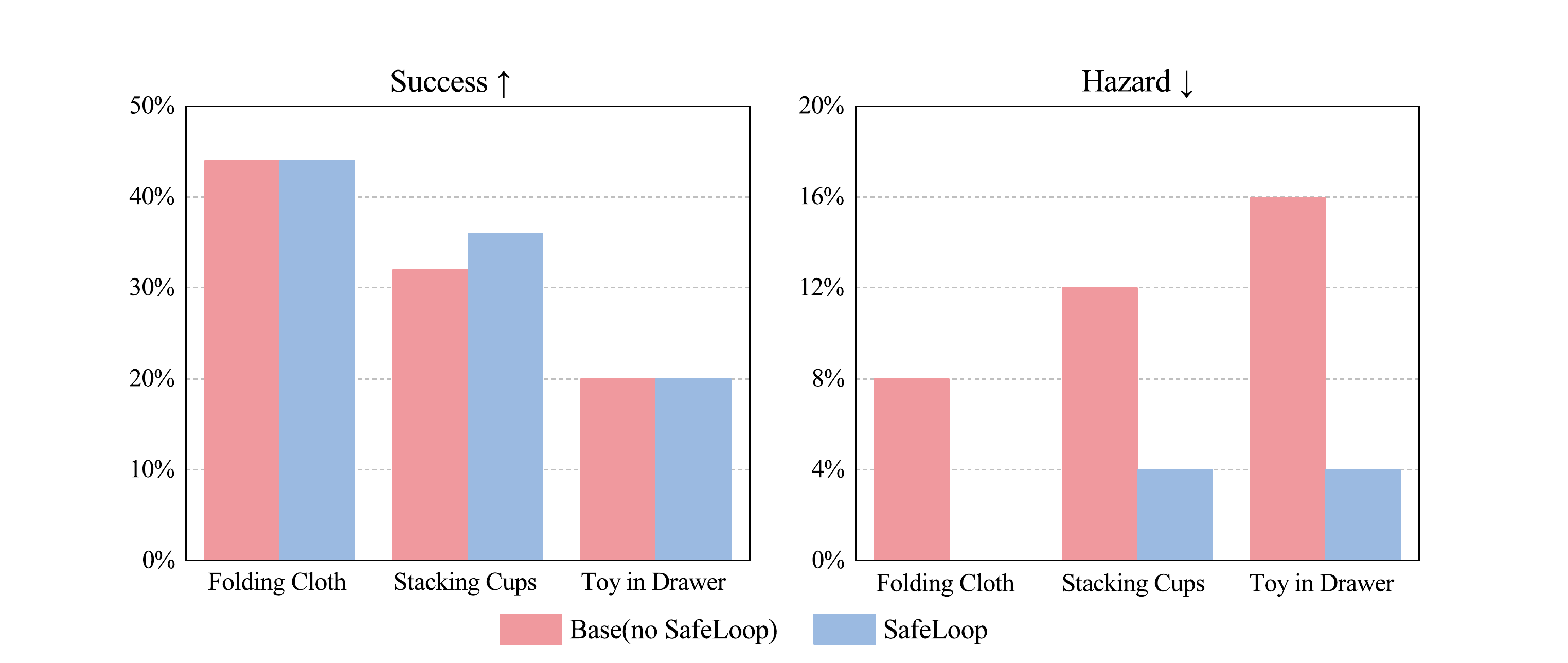}
  \caption{\textbf{Real-world validation.} Evaluation on three physical manipulation tasks (25 rollouts each). By shielding the simulation-trained decider from visual domain shifts, \textsc{SafeLoop} transfers directly to the real world, reducing hazard occurrences while preserving task success.}
  \label{fig:real_exp}
\end{figure}

\textbf{Sim-to-real safety transfer.}
We evaluate on three physical tasks: \textit{Folding Cloth}, \textit{Stacking Cups}, and \textit{Toy in Drawer} (Fig.~\ref{fig:real_setup}). The risk predictor is fine-tuned with labeled real-world images, whereas the RL decider is transferred unchanged from simulation; no real-world reinforcement learning is required. Compressing visual observations into the same four risk semantics limits the decider's exposure to visual domain shift. On average across the three tasks, \textsc{SafeLoop} reduces hazards by approximately 78\% while maintaining baseline success (Fig.~\ref{fig:real_exp}). After rollback to a safe waypoint, the stochastic base policy is queried again and may sample a continuation that avoids the developing hazard.

\section{Conclusion}
\label{sec:conclusion}

In this paper, we introduce \textsc{SafeLoop}, a non-invasive safety wrapper for frozen vision-language-action (VLA) policies. By decoupling short-horizon risk prediction from memory-based rollback, \textsc{SafeLoop} prevents developing hazards from becoming irreversible. It reduces simulated hazard-event rates by about 82\% across diverse VLA backbones (e.g., $1.21 \to 0.22$ Events/1k for \textsc{Pi-0}) while maintaining task success. Compressing visual observations into a compact risk vector enables the simulation-trained decider to transfer unchanged to physical robots, where adapting only the predictor reduces hazards by $\sim$78\% on average without real-world RL. \textsc{SafeLoop} does not update the base policy and therefore cannot guarantee a different continuation or undo irreversible scene changes, but it provides a modular execution-time safety mechanism for long-horizon manipulation.

\section*{Acknowledgment}
This study is supported by the Jiangsu Science and Technology Major Project (BG2025035) and the Fundamental Research Funds for the Central Universities (KG2025XX). This research is also supported by cash and in-kind funding from Nanjing Kunpeng\&Ascend Center of Cultivation and industry partner(s).

{
    \small
    \IEEEtriggeratref{29}
    \bibliographystyle{IEEEtran}
    \bibliography{main}

@STRING{IJRR = "Int. J. Robot. Res."}

@STRING{RAL  = "{IEEE} Robot. Autom. Lett."}

@STRING{NMI  = "Nat. Mach. Intell."}

@STRING{ICRA = "Proc. {IEEE} Int. Conf. Robotics and Automation ({ICRA})"}

@STRING{IROS = "Proc. {IEEE/RSJ} Int. Conf. Intelligent Robots and Systems ({IROS})"}

@STRING{CORL = "Proc. Conf. Robot Learning ({CoRL})"}

@STRING{RSS  = "Proc. Robotics: Science and Systems ({RSS})"}

@STRING{AAAI = "Proc. {AAAI} Conf. Artificial Intelligence"}

@article{brohan2022rt,
  title   = "{RT}-1: Robotics Transformer for Real-World Control at Scale",
  author  = "Anthony Brohan and Noah Brown and Justice Carbajal and Yevgen Chebotar and Joseph Dabis and Chelsea Finn and Keerthana Gopalakrishnan and Karol Hausman and Alex Herzog and Jasmine Hsu and others",
  journal = "arXiv preprint arXiv:2212.06817",
  year    = "2022"
}

@misc{yang2025qwen3technicalreport,
      title={Qwen3 Technical Report},
      author={An Yang and Anfeng Li and Baosong Yang and Beichen Zhang and Binyuan Hui and Bo Zheng and Bowen Yu and Chang Gao and Chengen Huang and Chenxu Lv and others},
      year={2025}
}

@misc{grattafiori2024llama3herdmodels,
      title={The Llama 3 Herd of Models},
      author={Aaron Grattafiori and Abhimanyu Dubey and Abhinav Jauhri and Abhinav Pandey and Abhishek Kadian and Ahmad Al-Dahle and Aiesha Letman and Akhil Mathur and Alan Schelten and Alex Vaughan and others},
      year={2024}
}

@misc{beyer2024paligemmaversatile3bvlm,
      title={PaliGemma: A versatile 3B VLM for transfer},
      author={Lucas Beyer and Andreas Steiner and André Susano Pinto and Alexander Kolesnikov and Xiao Wang and Daniel Salz and Maxim Neumann and Ibrahim Alabdulmohsin and Michael Tschannen and Emanuele Bugliarello and others},
      year={2024}
}

@article{cen2025rynnvla,
  title={RynnVLA-002: A Unified Vision-Language-Action and World Model},
  author={Cen, Jun and Huang, Siteng and Yuan, Yuqian and Li, Kehan and Yuan, Hangjie and Yu, Chaohui and Hou, Bing and Jiang, Yuming and Guo, Jiayan and Li, Xin and Luo, Hao and Wang, Fan and Zhao, Deli and Chen, Han},
  journal={arXiv preprint arXiv:2511.17502},
  year={2025}
}

@inproceedings{Ze2024DP3,
	title={3D Diffusion Policy: Generalizable Visuomotor Policy Learning via Simple 3D Representations},
	author={Yanjie Ze and Gu Zhang and Kangning Zhang and Chenyuan Hu and Muhan Wang and Huazhe Xu},
	booktitle={Proceedings of Robotics: Science and Systems (RSS)},
	year={2024}
}

@inproceedings{zhao2023learning,
  title={Learning fine-grained bimanual manipulation with low-cost hardware},
  author={Zhao, Tony Z and Kumar, Vikash and Levine, Sergey and Finn, Chelsea},
  booktitle={Proceedings of Robotics: Science and Systems},
  address={Daegu, Republic of Korea},
  month={July},
  year={2023}
}

@inproceedings{zitkovich2023rt,
  title       = "{RT}-2: Vision-Language-Action Models Transfer Web Knowledge to Robotic Control",
  author      = "Brianna Zitkovich and Tianhe Yu and Sichun Xu and Peng Xu and Ted Xiao and Fei Xia and Jialin Wu and Paul Wohlhart and Stefan Welker and Ayzaan Wahid and others",
  booktitle   = CORL,
  pages       = "2165--2183",
  year        = "2023",
  organization= "PMLR"
}

@article{black2024pi_0,
  title={$\pi _ {0} $: A Vision-Language-Action Flow Model for General Robot Control},
  author={Black, Kevin and Brown, Noah and Driess, Danny and Esmail, Adnan and Equi, Michael and Finn, Chelsea and Fusai, Niccolo and Groom, Lachy and Hausman, Karol and Ichter, Brian and others},
  journal={arXiv preprint arXiv:2410.24164},
  year={2024}
}

@article{chi2025diffusion,
  title   = {Diffusion Policy: Visuomotor Policy Learning via Action Diffusion},
  author  = {Chi, Cheng and Xu, Zhenjia and Feng, Siyuan and Cousineau, Eric and Du, Yilun and Burchfiel, Benjamin and Tedrake, Russ and Song, Shuran},
  journal = {The International Journal of Robotics Research},
  year    = {2025},
  doi     = {10.1177/02783649241273668}
}

@article{kim2024openvla,
  title   = "{OpenVLA}: An Open-Source Vision-Language-Action Model",
  author  = "Moo Jin Kim and Karl Pertsch and Siddharth Karamcheti and Ted Xiao and Ashwin Balakrishna and Suraj Nair and Rafael Rafailov and Ethan Foster and Grace Lam and Pannag Sanketi and others",
  journal = "arXiv preprint arXiv:2406.09246",
  year    = "2024"
}

@article{intelligence2025pi_,
  title   = "$\pi _ {0.5} $: A Vision-Language-Action Model with Open-World Generalization",
  author  = "{Physical Intelligence} and Kevin Black and Noah Brown and James Darpinian and Karan Dhabalia and Danny Driess and Adnan Esmail and Michael Equi and Chelsea Finn and Niccolo Fusai and others",
  journal = "arXiv preprint arXiv:2504.16054",
  year    = "2025"
}

@article{cen2025worldvla,
  title={{WorldVLA}: Towards autoregressive action world model},
  author={Cen, Jun and Yu, Chaohui and Yuan, Hangjie and Jiang, Yuming and Huang, Siteng and Guo, Jiayan and Li, Xin and Song, Yibing and Luo, Hao and Wang, Fan and Zhao, Deli and Chen, Han},
  journal={arXiv preprint arXiv:2506.21539},
  year={2025}
}

@article{ak2023learning,
  title     = "Learning Failure Prevention Skills for Safe Robot Manipulation",
  author    = "Abdullah Cihan Ak and Eren Erdal Aksoy and Sanem Sariel",
  journal   = RAL,
  volume    = "8",
  number    = "12",
  pages     = "7994--8001",
  year      = "2023",
  publisher = "{IEEE}"
}

@inproceedings{thumm2022provably,
  title       = "Provably Safe Deep Reinforcement Learning for Robotic Manipulation in Human Environments",
  author      = "Jakob Thumm and Matthias Althoff",
  booktitle   = ICRA,
  pages       = "6344--6350",
  year        = "2022",
  organization= "{IEEE}"
}

@article{deng2025safebimanual,
  title={SafeBimanual: Diffusion-based trajectory optimization for safe bimanual manipulation},
  author={Deng, Haoyuan and Guo, Wenkai and Wang, Qianzhun and Wu, Zhenyu and Wang, Ziwei},
  journal={arXiv preprint arXiv:2508.18268},
  year={2025}
}

@article{kim2025openvla_oft,
  title   = "{Fine-Tuning Vision-Language-Action Models: Optimizing Speed and Success}",
  author  = "Moo Jin Kim and Chelsea Finn and Percy Liang",
  journal = "arXiv preprint arXiv:2502.19645",
  year    = "2025"
}

@inproceedings{amaya2022vision,
  title       = "Vision-Based Safety System for Barrierless Human-Robot Collaboration",
  author      = "Lina Mar{\'\i}a Amaya-Mej{\'\i}a and Nicol{\'a}s Duque-Su{\'a}rez and Daniel Jaramillo-Ram{\'\i}rez and Carol Martinez",
  booktitle   = IROS,
  pages       = "7331--7336",
  year        = "2022",
  organization= "{IEEE}"
}

@article{brunke2025semantically,
  title   = {Semantically Safe Robot Manipulation: From Semantic Scene Understanding to Motion Safeguards},
  author  = {Brunke, Lukas and Zhang, Yanni and R{\"o}mer, Ralf and Naimer, Jack and Staykov, Nikola and Zhou, Siqi and Schoellig, Angela P.},
  journal = {IEEE Robotics and Automation Letters},
  year    = {2025},
  volume  = {10},
  number  = {5},
  pages   = {4810--4817}
}

@article{khatib1986real,
  title     = "Real-Time Obstacle Avoidance for Manipulators and Mobile Robots",
  author    = "Oussama Khatib",
  journal   = IJRR,
  volume    = "5",
  number    = "1",
  pages     = "90--98",
  year      = "1986",
  publisher = "SAGE Publications"
}

@article{zucker2013chomp,
  title     = "{CHOMP}: Covariant Hamiltonian Optimization for Motion Planning",
  author    = "Matt Zucker and Nathan Ratliff and Anca D. Dragan and Mihail Pivtoraiko and Matthew Klingensmith and Christopher M. Dellin and J. Andrew Bagnell and Siddhartha S. Srinivasa",
  journal   = IJRR,
  volume    = "32",
  number    = "9-10",
  pages     = "1164--1193",
  year      = "2013",
  publisher = "SAGE Publications"
}

@inproceedings{schulman2013finding,
  title     = "Finding Locally Optimal, Collision-Free Trajectories with Sequential Convex Optimization",
  author    = "John Schulman and Jonathan Ho and Alex X. Lee and Ibrahim Awwal and Henry Bradlow and Pieter Abbeel",
  booktitle = RSS,
  year      = "2013"
}

@article{nubert2020safe,
  title     = "Safe and Fast Tracking on a Robot Manipulator: Robust {MPC} and Neural Network Control",
  author    = "Julian Nubert and Johannes K{\"o}hler and Vincent Berenz and Frank Allg{\"o}wer and Sebastian Trimpe",
  journal   = RAL,
  volume    = "5",
  number    = "2",
  pages     = "3050--3057",
  year      = "2020",
  publisher = "{IEEE}"
}

@inproceedings{murtaza2021real,
  title       = "Real-Time Safety and Control of Robotic Manipulators with Torque Saturation in Operational Space",
  author      = "Muhammad Ali Murtaza and Sergio Aguilera and Vahid Azimi and Seth Hutchinson",
  booktitle   = IROS,
  pages       = "702--708",
  year        = "2021",
  organization= "{IEEE}"
}

@inproceedings{alshiekh2018shielding,
  title     = {Safe Reinforcement Learning via Shielding},
  author    = {Alshiekh, Mohammed and Bloem, Roderick and Ehlers, R{\"u}diger and K{\"o}nighofer, Bettina and Niekum, Scott and Topcu, Ufuk},
  booktitle = {Proc. {AAAI} Conf. Artificial Intelligence},
  year      = {2018},
  pages     = {2669--2678}
}

@article{dalal2018safe,
  title   = "Safe Exploration in Continuous Action Spaces",
  author  = "Gal Dalal and Krishnamurthy Dvijotham and Matej Vecerik and Todd Hester and Cosmin Paduraru and Yuval Tassa",
  journal = "arXiv preprint arXiv:1801.08757",
  year    = "2018",
  url     = "https://arxiv.org/abs/1801.08757"
}

@article{intelligence2025pi,
  title={$\pi^{*} _ {0.6} $: a VLA That Learns From Experience},
  author={{Physical Intelligence} and Amin, Ali and Aniceto, Raichelle and Balakrishna, Ashwin and Black, Kevin and Conley, Ken and Connors, Grace and Darpinian, James and Dhabalia, Karan and DiCarlo, Jared and others},
  journal={arXiv preprint arXiv:2511.14759},
  year={2025}
}

@misc{gr00tn1_2025,
  title      = {{GR00T} {N1}: An Open Foundation Model for Generalist Humanoid Robots},
  author     = {{NVIDIA} and Johan Bjorck and Fernando Castañeda and Nikita Cherniadev and Xingye Da and Runyu Ding and Linxi J. Fan and Yu Fang and Dieter Fox and Fengyuan Hu and Spencer Huang and Joel Jang and Zhenyu Jiang and Jan Kautz and Kaushil Kundalia and Lawrence Lao and Zhiqi Li and Zongyu Lin and Kevin Lin and Guilin Liu and Edith Llontop and Loic Magne and Ajay Mandlekar and Avnish Narayan and Soroush Nasiriany and Scott Reed and You Liang Tan and Guanzhi Wang and Zu Wang and Jing Wang and Qi Wang and Jiannan Xiang and Yuqi Xie and Yinzhen Xu and Zhenjia Xu and Seonghyeon Ye and Zhiding Yu and Ao Zhang and Hao Zhang and Yizhou Zhao and Ruijie Zheng and Yuke Zhu},
  month      = {March},
  year       = {2025}
}

@article{thananjeyan2021recovery,
  title     = "{RecoveryRL}: Safe Reinforcement Learning with Learned Recovery Zones",
  author    = "Brijen Thananjeyan and Ashwin Balakrishna and Suraj Nair and Michael Luo and Krishnan Srinivasan and Minho Hwang and Joseph E. Gonzalez and Julian Ibarz and Chelsea Finn and Ken Goldberg",
  journal   = RAL,
  volume    = "6",
  number    = "3",
  pages     = "4915--4922",
  year      = "2021",
  publisher = "{IEEE}"
}

@inproceedings{zhang2025safevla,
  title     = {SafeVLA: Towards Safety Alignment of Vision-Language-Action Model via Constrained Learning},
  author    = {Zhang, Borong and Zhang, Yuhao and Ji, Jiaming and Lei, Yingshan and Dai, Josef and Chen, Yuanpei and Yang, Yaodong},
  booktitle = {Advances in Neural Information Processing Systems (NeurIPS)},
  year      = {2025},
  note      = {Spotlight Presentation}
}

@inproceedings{xue2025reactive,
  title     = {Reactive Diffusion Policy: Slow-Fast Visual-Tactile Policy Learning for Contact-Rich Manipulation},
  author    = {Xue, Han and Ren, Jieji and Chen, Wendi and Zhang, Gu and Fang, Yuan and Gu, Guoying and Xu, Huazhe and Lu, Cewu},
  booktitle = {Proceedings of Robotics: Science and Systems (RSS)},
  year      = {2025}
}

@inproceedings{NEURIPS2024_ca92ff06,
 author = {Chen, Zixuan and Ji, Ze and Huo, Jing and Gao, Yang},
 booktitle = {Advances in Neural Information Processing Systems},
 doi = {10.52202/079017-3547},
 editor = {A. Globerson and L. Mackey and D. Belgrave and A. Fan and U. Paquet and J. Tomczak and C. Zhang},
 pages = {111679--111714},
 publisher = {Curran Associates, Inc.},
 title = {SCaR: Refining Skill Chaining for Long-Horizon Robotic Manipulation via Dual Regularization},
 url = {https://proceedings.neurips.cc/paper_files/paper/2024/file/ca92ff06d973ece92cecc561757d500e-Paper-Conference.pdf},
 volume = {37},
 year = {2024}
}

@article{schulman2017proximal,
  title   = {Proximal Policy Optimization Algorithms},
  author  = {Schulman, John and Wolski, Filip and Dhariwal, Prafulla and Radford, Alec and Klimov, Oleg},
  journal = {arXiv preprint arXiv:1707.06347},
  year    = {2017}
}

@inproceedings{gu2025safe,
  title     = {SAFE: Multitask Failure Detection for Vision-Language-Action Models},
  author    = {Gu, Qiao and Ju, Yuanliang and Sun, Shengxiang and Gilitschenski, Igor and Nishimura, Haruki and Itkina, Masha and Shkurti, Florian},
  booktitle = {Advances in Neural Information Processing Systems (NeurIPS)},
  year      = {2025}
}

@article{hu2025vlsa,
  title={{VLSA}: Vision-Language-Action Models with Plug-and-Play Safety Constraint Layer},
  author={Hu, Songqiao and Liu, Zeyi and Liu, Shuang and Cen, Jun and Meng, Zihan and Wang, Sheng and Li, Xin and He, Xiao},
  journal={arXiv preprint arXiv:2512.11891},
  year={2025}
}

@inproceedings{hsu2022improving,
  title       = "Improving Safety in Deep Reinforcement Learning Using Unsupervised Action Planning",
  author      = "Hao-Lun Hsu and Qiuhua Huang and Sehoon Ha",
  booktitle   = ICRA,
  pages       = "5567--5573",
  year        = "2022",
  organization= "{IEEE}"
}

@inproceedings{farid2022failure,
  title   = "Failure Prediction with Statistical Guarantees for Vision-Based Robot Control",
  author  = "Alec Farid and David Snyder and Allen Z. Ren and Anirudha Majumdar",
  booktitle = "Proceedings of Robotics: Science and Systems",
  address = "New York City, NY, USA",
  month = jun,
  year    = "2022"
}

@article{yang2025agentic,
  title   = "Agentic Robot: A Brain-Inspired Framework for Vision-Language-Action Models in Embodied Agents",
  author  = "Zhejian Yang and Yongchao Chen and Xueyang Zhou and Jiangyue Yan and Dingjie Song and Yinuo Liu and Yuting Li and Yu Zhang and Pan Zhou and Hechang Chen and Lu Sun",
  journal = "arXiv preprint arXiv:2505.23450",
  year    = "2025"
}

@article{lbmtri2025,
  title={A Careful Examination of Large Behavior Models for Multitask Dexterous Manipulation},
  author={{TRI LBM Team} and Jose Barreiros and Andrew Beaulieu and Aditya Bhat and Rick Cory and Eric Cousineau and Hongkai Dai and Ching-Hsin Fang and Kunimatsu Hashimoto and Muhammad Zubair Irshad and others},
  journal={arXiv preprint arXiv:2507.05331},
  year={2025},
  url={https://arxiv.org/abs/2507.05331}
}

@inproceedings{liu2023libero,
  title={LIBERO: Benchmarking Knowledge Transfer for Lifelong Robot Learning},
  author={Liu, Bo and Zhu, Yifeng and Gao, Chongkai and Feng, Yihao and Liu, Qiang and Zhu, Yuke and Stone, Peter},
  booktitle={Advances in Neural Information Processing Systems},
  volume={36},
  year={2023}
}

@article{triantafyllidis2023hybrid,
  title     = "Hybrid Hierarchical Learning for Solving Complex Sequential Tasks Using the Robotic Manipulation Network {ROMAN}",
  author    = "Eleftherios Triantafyllidis and Fernando Acero and Zhaocheng Liu and Zhibin Li",
  journal   = NMI,
  volume    = "5",
  number    = "9",
  pages     = "991--1005",
  year      = "2023",
  publisher = "Nature Publishing Group"
}

@article{vats2024recoverychaining,
  title   = "{RecoveryChaining}: Learning Local Recovery Policies for Robust Manipulation",
  author  = "Shivam Vats and Devesh K. Jha and Maxim Likhachev and Oliver Kroemer and Diego Romeres",
  journal = "arXiv preprint arXiv:2410.13979",
  year    = "2024"
}

@inproceedings{kim2023machine,
  title     = "A Machine with Short-Term, Episodic, and Semantic Memory Systems",
  author    = "Taewoon Kim and Michael Cochez and Vincent Fran{\c{c}}ois-Lavet and Mark Neerincx and Piek Vossen",
  booktitle = AAAI,
  volume    = "37",
  number    = "1",
  pages     = "48--56",
  year      = "2023"
}
}
\end{document}